%% file: main.tex
\documentclass[10pt,twocolumn,letterpaper]{article}

\usepackage{cvpr}              % To produce the CAMERA-READY version

\usepackage{booktabs,tabularx,xcolor,amssymb}
\usepackage{pifont}
\newcommand{\cmark}{\ding{51}}  % checkmark
\newcommand{\xmark}{\ding{55}}  % cross
\newcommand{\pmark}{\(\sim\)}     % partial
\def \eg {\textit{e.g.}}
\def \ie {\textit{i.e.}}

\usepackage{graphicx}
\usepackage[table]{xcolor}  % 支持表格颜色
\usepackage{colortbl}       % 支持表格颜色（更稳定）
\usepackage{multirow}
\usepackage{capt-of}
\usepackage{listings}
\usepackage{xcolor}
\newcounter{magicrownumbers}

\newcommand \footnoteONLYtext[1]
{
    \let \mybackup \thefootnote
    \let \thefootnote \relax
    \footnotetext{#1}
    \let \thefootnote \mybackup
    \let \mybackup \imareallyundefinedcommand
}

\input{preamble}
\definecolor{cvprblue}{rgb}{0.21,0.49,0.74}
\usepackage[pagebackref,breaklinks,colorlinks,allcolors=cvprblue]{hyperref}

\def\paperID{2296} % *** Enter the Paper ID here
\def\confName{CVPR}
\def\confYear{2026}

\title{SurgCoT: Advancing Spatiotemporal Reasoning in Surgical Videos through a Chain-of-Thought Benchmark} 

\author{
Gui Wang$^{1,2}$,~
YongSong Zhou$^{1}$,~
Kaijun Deng$^{1}$,~
Wooi Ping Cheah$^{2}$,~
Rong Qu$^{2}$,\\
Jianfeng Ren$^{*,2}$,~
Linlin Shen$^{*,3,4}$~\\
\textsuperscript{\rm1}School of Computer Science and Software Engineering, Shenzhen University\\
\textsuperscript{\rm2}University of Nottingham \quad \textsuperscript{\rm3}School of AI, Shenzhen University\\
\textsuperscript{\rm4}Guangdong Provincial Key Laboratory of Intelligent Information Processing, Shenzhen University\\
{\tt\small jianfeng.ren@nottingham.edu.cn;~llshen@szu.edu.cn}
}

\begin{document}
% \maketitle

\twocolumn[{
\renewcommand\twocolumn[1][]{#1}
\maketitle
\begin{center}
    \vspace{-16pt}
    \centering
    \captionsetup{type=figure}
    \includegraphics[width=0.94\linewidth]{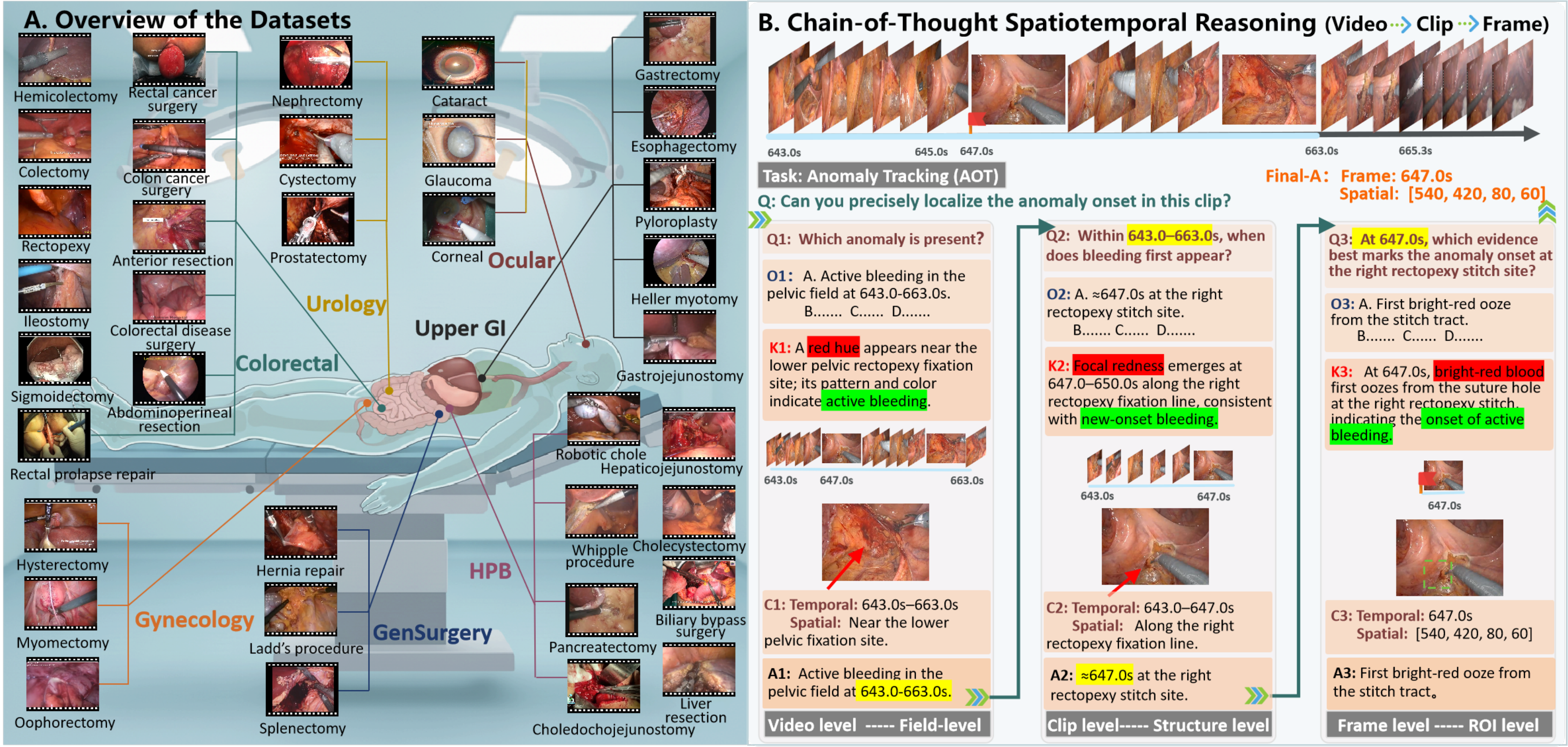}
        \caption{SurgCoT comprises \textbf{2,841 surgical videos} across \textbf{7 specialties} and \textbf{35 procedures}, with \textbf{19,345 main questions} and \textbf{59,177 sub-questions}.
        SurgCoT advances beyond frame-level tasks (\eg, phase/tool recognition) by introducing a three-stage, five-tuple annotation protocol (\texttt{Question→Option→Knowledge→Clue→Answer}) to scaffold chain-of-thought reasoning. The framework’s efficacy stems from its multi-stage reasoning structure and the synergistic interaction between the \texttt{Knowledge} field, which supplies contextual background, and the \texttt{Clue} field, which provides targeted spatiotemporal evidence, jointly enabling hierarchical reasoning.          
        % Moving beyond frame-level tasks (\eg, phase/tool recognition), SurgCoT introduces a novel three-stage, five-tuple annotation protocol (\texttt{Question→Option→Knowledge→Clue→Answer}) to scaffold a chain-of-thought reasoning process. Its efficacy centers on not only the three-stage progressive reasoning chains, but also the synergy between the \texttt{Knowledge} field, which provides essential background context, and the \texttt{Clue} field, which supplies targeted spatiotemporal evidence, thereby scaffolding hierarchical reasoning. 
        %\rjf{The current drawing can't clearly show the benefits of "think" and "clue", e.g., articulates the reasoning chain and supplies targeted spatiotemporal hints to navigate it. Suggest to remove previous work and focus on how think and clue work.}
        }  
        % Unlike prior frame-level tasks (\eg, phase/tool recognition), it introduces a three-step, five-tuple annotation protocol (\texttt{Question→Option→Think→Clue→Answer}), where  where the \texttt{Think} field explicitly models the reasoning process by decomposing the problem into sub-questions, and the \texttt{Clue} field provides progressive spatiotemporal hints to guide models through hierarchical reasoning and localization.} 
    % \caption{Overview of SurgBench-5ST. \textbf{Left}: Example of Surgical videos list, including 2003 videos, 7 surgical specialties, and 35 procedures. \textbf{Right}: Details of previous benchmarks and our anomaly onset tracking example. Each sample follows a \textbf{three-step} (Q1$\rightarrow$Q2$\rightarrow$Q3) \textbf{five-tuple} annotation protocol, \emph{question/option/think/clue/answer}, where the \texttt{clue} progressively narrows the temporal and spatial scope to guide hierarchical reasoning and localization.}
    \label{overview}
    \vspace{-2pt}
\end{center}
}]

\footnoteONLYtext{*Corresponding author}
\input{sec/0_abstract}    
\input{sec/1_intro}

\input{sec/2_relatedwork}
\input{sec/3_SurgBench5ST}
\input{sec/4_Experiments}

\input{sec/5_conclusion}

\clearpage
\newpage
\section*{Acknowledgement}
This work was supported by the National Natural Science Foundation of China (Grant No. 62576216), Ningbo Municipal Bureau of Science and Technology (Grant Nos. 2023Z138, 2023Z237, 2024Z110, and 2024Z124), and the Guangdong Provincial Key Laboratory (Grant No. 2023B1212060076). The work was also supported by the Intelligent Computing Center of Shenzhen University.

{
    \small
    \bibliographystyle{ieeenat_fullname}
    \bibliography{main}
}


\end{document}

%% file: sec/0_abstract.tex
\begin{abstract}
Fine-grained spatiotemporal reasoning on surgical videos is critical, yet the capabilities of Multi-modal Large Language Models (MLLMs) in this domain remain largely unexplored. To bridge this gap, we introduce \textbf{SurgCoT}, a unified benchmark for evaluating chain-of-thought (CoT) reasoning in MLLMs across \textbf{7 surgical specialties} and \textbf{35 diverse procedures}. SurgCoT assesses five core reasoning dimensions: Causal Action Ordering, Cue–Action Alignment, Affordance Mapping, Micro‑Transition Localization, and Anomaly Onset Tracking, through a structured CoT framework with an intensive annotation protocol (\texttt{Question→Option→Knowledge→Clue→Answer}), where the \texttt{Knowledge} field provides essential background context and \texttt{Clue} provides definitive spatiotemporal evidence. Evaluation of 10 leading MLLMs shows: 1) commercial models outperform open-source and medical-specialized variants; 2) significant gaps exist in surgical CoT reasoning; 3) SurgCoT enables effective evaluation and enhances progressive spatiotemporal reasoning. SurgCoT provides a reproducible testbed to narrow the gap between MLLM capabilities and clinical reasoning demands. Code: https://github.com/CVI-SZU/SurgCoT. 
\end{abstract}

%% file: sec/1_intro.tex
\section{Introduction}
\label{sec:intro}
Surgical videos are a cornerstone of perioperative care and education, rich with dynamic anatomical and procedural data \cite{yip2023artificial,maier2022surgical}. The growing volume and complexity of this data make fine-grained spatiotemporal understanding critical for clinical safety and efficiency \cite{lam2022machine}. In recent years, Multi-modal Large Language Models (MLLMs) have been adapted for surgical settings, enabling interactive analysis for tasks like surgical phase recognition~\cite{PSI-AVA-VQA,zhang2025mejo,Surg-396K}, instrument recognition~\cite{huang2025mllms,chen2025surgllm}, tissue detection~\cite{huang2025mllms,chen2025surgllm}, and surgery understanding~\cite{EndoAssistant,chen2024vs,hao2025surgery}. This rapid progress, however, creates a parallel need for robust evaluation frameworks to assess their true clinical utility.

% Surgical videos serve as a core medium for perioperative care and surgical education, encapsulating rich anatomical semantics, evolving operative fields, and instrument–tissue interactions~\cite{yip2023artificial,maier2022surgical}. As the volume and complexity of surgical video data continue to grow, achieving efficient and reliable fine-grained spatiotemporal understanding has become increasingly critical for clinical safety and operational efficiency~\cite{lam2022machine}. In recent years, Multi-modal Large Language Models (MLLMs) have made notable progress on general tasks, prompting a wave of explorations tailored to medical and surgical settings~\cite{PSI-AVA-VQA,huang2025mllms,hao2025surgery,chen2024vs,chen2025surgllm,Surg-396K,zhang2025mejo}. These models enable interactive analysis of surgical videos via text prompts and support a range of clinical tasks, including surgical phase recognition~\cite{PSI-AVA-VQA,zhang2025mejo,Surg-396K}, instrument recognition~\cite{huang2025mllms,chen2025surgllm}, tissue detection~\cite{huang2025mllms,chen2025surgllm}, and surgery understanding~\cite{EndoAssistant,chen2024vs,hao2025surgery}. With the emergence of such surgery-oriented MLLMs, there is a parallel need for robust evaluation frameworks to assess their clinical utility and performance.

Surgical benchmarks fall into two categories: 
General-purpose benchmarks~\cite{SSG-VQA,PSI-AVA-VQA}, covering broad scenarios but emphasize generic question answering (\eg, phases, instruments) while specialized benchmarks~\cite{SurgQA,SurVLP} focus on narrow procedural domains (\eg, ophthalmic, endoscopic) but rely on frame/clip-level VQAs that treat videos as discrete segments, ignoring cross-temporal dependencies and failing to evaluate spatiotemporal or causal reasoning~\cite{MedFrameQA,SFN-ESVQA}. In contrast, surgeons must track subtle, rapid spatiotemporal changes to support fine-grained inference and decision-making~\cite{liu2025lovit, szydlo2025surgical}. A critical question remains: \textit{Can MLLMs achieve expert-level progressive spatiotemporal reasoning and causal judgment across diverse surgical contexts?}

\begin{table*}[t]
\caption{Comparison of surgical benchmarks. Our SurgCoT uniquely spans 7 surgical specialties with multi-level annotations (video/clip/frame), supporting hierarchical spatiotemporal reasoning with localization supervision and clinician-derived reference standards. 
}
\label{tab:bench-compact}
\centering
% \scriptsize
\small
\begin{tabularx}{\textwidth}{lcccccccccc}
\toprule
\textbf{Benchmark} & \textbf{Domain}  & \textbf{\#Spe.} & \textbf{\#Pro.} & \textbf{Scale} & \textbf{Unit} & \textbf{ST} & \textbf{Pro.}  & \textbf{Loc.} & \textbf{Clin.} \\
\midrule
Surgical-VQA~\cite{Surgical-vqa}$_{\text{(MICCAI'21)}}$ & Endoscope & \pmark & 3 & 11.8K & Frame & \xmark & \xmark & \cmark & \cmark\\
EndoVis-VQLA~\cite{EndoVis-VQLA}$_{\text{(ICRA'23)}}$ & Endoscope & \pmark & 1 & 9.5K & Frame & \xmark & \xmark & \cmark & \cmark\\
Endo-FM~\cite{Endo-FM}$_{\text{(MICCAI'23)}}$ & Endoscope & \pmark & 2 & 33K+50K & Clip/Frame & \xmark & \xmark & \cmark & \cmark\\
CoPESD~\cite{CoPESD}$_{\text{(arXiv'24)}}$ & Endoscope & \pmark & 1 & 17.7K & Frame & \xmark & \cmark & \cmark & \cmark\\
SurVLP~\cite{SurVLP}$_{\text{(MedIA'25)}}$ & Endoscope & \pmark & 11 & 25.5K & Frame & \xmark & \xmark & \xmark & \xmark\\
Surg-3M~\cite{Surg-3M}$_{\text{(arXiv'25)}}$ & Endoscope & \pmark & 35 & 30K & Frame & \xmark & \xmark & \xmark & \xmark\\
Surg-396K~\cite{Surg-396K}$_{\text{(MIA'26)}}$ & Endoscope & 3 & 3 & 396K & Frame & \xmark & \xmark & \cmark & \cmark\\
\midrule
Cholec80-VQA~\cite{Surgical-vqa}$_{\text{(MICCAI'22)}}$ & Laparoscopic & 1 & 1 & 43K & Frame & \xmark & \xmark & \xmark & \cmark\\
GSViT~\cite{GSViT}$_{\text{(arXiv'24)}}$ & Laparoscopic & 1 & 28 & 7000K & Frame & \xmark & \xmark & \xmark & \xmark \\
GenSurg+~\cite{GenSurg+}$_{\text{(arXiv'24)}}$ & Laparoscopic  & 2 & 28  & 17K & Clip & \xmark & \xmark & \xmark & \xmark \\
\midrule
MedVidQA~\cite{MedVidQA}$_{\text{(BioNLP@ACL'24)}}$  & Instructional  & \pmark & \pmark & 1.2K & Clip & \xmark & \xmark & \xmark & \cmark \\
Surg-QA~\cite{SurgQA}$_{\text{(arXiv'24)}}$  & Instructional  & \pmark & \pmark & 10.2K & Video & \xmark & \xmark & \xmark & \xmark \\
M $^{3}$-Med~\cite{M3-Med}$_{\text{(arXiv'25)}}$  & Instructional  & \pmark & \pmark & 3.7K & Clip & \xmark & \xmark & \xmark & \cmark \\
\midrule
OphNet~\cite{ophnet}$_{\text{(ECCV'24)}}$ & Ophthalmic & 1 & \pmark & 14K & Video & \xmark & \cmark & \cmark & \cmark\\
OphVL~\cite{hu2024ophclip}$_{\text{(arXiv'24)}}$ & Ophthalmic & 1 & \pmark & 375K & Clip & \xmark & \xmark & \xmark & \cmark\\
EyePCR~\cite{EyePCR}$_{\text{((arXiv'25)}}$ & Ophthalmic & 1 & \pmark & 210K & Clip/Frame & \xmark & \cmark & \xmark & \cmark\\
\midrule
PSI-AVA-VQA~\cite{PSI-AVA-VQA}$_{\text{(MICCAI'23)}}$ & Generalist & 1 & 1 & 10.3K & Frame & \xmark & \xmark & \cmark & \cmark\\
SSG-VQA~\cite{SSG-VQA}$_{\text{(IJCARS'24)}}$ & Generalist & 1 & 1 & 960K & Frame & \xmark & \xmark & \cmark & \cmark\\
SurgVLM-Bench~\cite{SurgVLM}$_{\text{(arXiv'25)}}$ & Generalist & 4 & 16 & 7798.4K & Frame & \xmark & \cmark & \cmark & \xmark\\
MedFrameQA~\cite{MedFrameQA}$_{\text{(arXiv'25)}}$ & Generalist & 9 & \pmark & 3.4K & Multi-Frame & \cmark & \xmark & \xmark & \cmark\\
SurgBench~\cite{SurgBench}$_{\text{(arXiv'25)}}$ & Generalist & 11 & 22 & 23K+5300K & Clip/Frame & \xmark & \xmark & \cmark & \cmark\\
\rowcolor{gray!12}
\textbf{SurgCoT$_{\text{(Ours)}}$} & \textbf{Generalist} & \textbf{7} & \textbf{35} & 2.8K+19K+19K & Video/Clip/Frame & \cmark & \cmark & \cmark & \cmark\\
\bottomrule
\end{tabularx}
\vspace{2pt}
\footnotesize \textbf{\#Spe.}: \#Specialties; \textbf{\#Pro.}:\#Procedures; \textbf{ST}: Spatiotemporal modeling; \textbf{Pro.}: Progressive reasoning protocol; \textbf{Loc.}: Localization supervision (caption/bbox/mask/timestamps/tracks.); \textbf{Clin.}: Clinicians as a reference standard. \cmark: yes, \xmark: no, \pmark: unknown.
\end{table*}

To tackle this, we introduce \textbf{SurgCoT} (Fig.~\ref{overview}), a unified surgical video benchmark for evaluating MLLMs under a chain-of-thought (CoT) reasoning framework. SurgCoT advances the field in two key aspects:
1)~\textbf{Comprehensive Task Coverage}: Unlike prior benchmarks focused on narrow tasks~\cite{EyePCR,SurgQA,SurVLP} or generic queries~\cite{SSG-VQA,PSI-AVA-VQA,MedFrameQA}, SurgCoT spans 35 procedures across 7 diverse specialties, 
enabling broad and realistic assessment of expert-level cognitive reasoning. 
2)~\textbf{In-depth Reasoning Dimensions}: SurgCoT introduces 5 clinical-relevant reasoning dimensions (\textit{Causal Action Ordering}, \textit{Cue-Action Alignment}, \textit{Affordance Mapping}, \textit{Micro-Transition Localization}, and \textit{Anomaly Onset \& Tracking}), reflecting critical cognitive processes in real world surgery. Collectively, these rich reasoning dimensions on diverse procedures enable a fine-grained evaluation of spatiotemporal understanding and clinical decision-making, closely mirroring actual surgical workflows.

To enable CoT reasoning in surgical video understanding, as shown in Fig.~\ref{overview}, we introduce a three-stage progressive reasoning framework (\texttt{Q1→Q2→Q3}) with a comprehensive five-tuple annotation protocol (\texttt{Question→Option→Knowledge→Clue→Answer}). The novelty is two-fold: 1) It establishes a hierarchical reasoning pathway progressing from video-level comprehension through clip-level analysis to frame-level localization. As shown in Fig.~\ref{overview}, all five annotation tuples evolve through these stages, with each stage's \texttt{Answer} serving as contextual evidence for the subsequent stage, forming a continuous reasoning chain. 
2) It employs synergistic annotation strategies where the \texttt{Knowledge} field provides clinically relevant background context based on the \texttt{Question} and \texttt{Option}, while the \texttt{Clue} supplies targeted spatiotemporal evidence extracted directly from the surgical video content. 
This integrated design progressively narrows both temporal and spatial scope, effectively guiding MLLMs through structured hierarchical reasoning. 

To validate its effectiveness as a unified spatiotemporal reasoning benchmark, we systematically evaluate 12 state-of-the-art MLLMs across diverse surgical procedures within a CoT framework, including open-source~\cite{xu2025qwen3, wang2024qwen2, llavavideo,llava_next_interleave_2024}, medically specialized ~\cite{sellergren2025medgemma, xu2025lingshu, chen2024huatuogpt}, and commercial variants~\cite{openai2025gpt5, claude_sonnet_4_5, google2024gemini}. Results show that while commercial models lead their counterparts, all models exhibit significant limitations in fine-grained spatiotemporal understanding. Notably, by utilizing the three-stage progressive reasoning framework with five-tuple annotation protocol, considerable accuracy improvements could be achieved, indicating a viable path toward enhanced reasoning capability. SurgCoT offers a reproducible benchmark for method development and fair evaluation in whole-body surgical video analysis, highlighting both recent progress and the remaining gap to clinical-level reasoning.

Our contributions are three-fold: 
1)~We introduce SurgCoT, the first surgical video reasoning benchmark offering cross-specialty procedural coverage and a five-dimensional evaluation framework for fine-grained assessment of spatiotemporal understanding and clinical decision-making aligned with surgical workflows. 
2)~We propose a novel three-stage reasoning framework with a five-tuple annotation protocol, combining hierarchical video-to-frame localization with synergistic \texttt{Knowledge}/\texttt{Clue} annotations to enable structured CoT reasoning.
3)~Our evaluation of 12 leading MLLMs reveals that significant gaps persist in surgical CoT reasoning, while SurgCoT effectively enables progressive spatiotemporal reasoning evaluation.

%% file: sec/2_relatedwork.tex
\section{Related Work}
\label{sec:Relatedwork}

\noindent\textbf{Multi-Modal Large Language Models.} \quad 
MLLMs have achieved notable performance on complex vision-language tasks through large-scale pretraining and instruction tuning. Initial joint-encoder models such as BLIP~\cite{li2022blip,li2023blip} and Flamingo~\cite{alayrac2022flamingo} connected visual encoders with cross-attention layers for multimodal fusion. Subsequent autoregressive MLLMs like LLaVA~\cite{liu2023visual} tokenize images as visual prompts for LLM conditioning, showing strong few-shot generalization. Recent models such as Qwen-VL~\cite{wang2024qwen2} and InternVL~\cite{chen2024internvl} scale visual backbones and alignment methods, approaching the performance of proprietary systems like GPT-4o~\cite{hurst2024gpt}. Video-language models such as Video-LLaVA~\cite{Video-llava} further improve temporal modeling, offering a suitable substrate for surgical video analysis.

In the medical domain, MLLM adaptation has progressed through specialized pretraining and instruction tuning, \eg, LLaVA-Med~\cite{li2023llava} leverage biomedical corpora to enhance visual diagnostic capabilities, while HuatuoGPT-Vision~\cite{chen2024huatuogpt} introduces large-scale medical VQA datasets for knowledge injection. Retrieval-augmented approaches such as MedDr~\cite{Meddr} further strengthen factual grounding through external knowledge bases. Surgical-specific MLLMs including SurgicalGPT~\cite{PSI-AVA-VQA} and Surgical-LVLM~\cite{Surgical-lvlm} have demonstrated promising results in operative tasks like phase recognition and report generation. However, current evaluations remain confined to narrow tasks and short video segments, leaving fine-grained spatiotemporal reasoning largely unexamined~\cite{SFN-ESVQA,MedFrameQA}. This limitation underscores the need for unified benchmarks capable of assessing expert-level reasoning in realistic surgical workflows.

\noindent\textbf{Benchmarks for Surgical MLLMs.} \quad 
Table~\ref{tab:bench-compact} summarizes the landscape of surgical benchmarks. Early works including SurgicalVQA and Cholec80-VQA~\cite{Surgical-vqa} establish foundational protocols, while subsequent efforts like SSG-VQA~\cite{SSG-VQA} extend to complex laparoscopic scenarios with enhanced geometric and procedural analysis. Specialized benchmarks (\eg, EndoVis-VQLA~\cite{EndoVis-VQLA} and CoPESD~\cite{CoPESD}) focus on tool analysis and localization within narrow procedural contexts, but with limited generalization. 
To address single-specialty constraints, cross-specialty benchmarks have emerged: SurgVLM-Bench~\cite{SurgVLM} covers 16 procedures across 4 specialties with hierarchical knowledge organization, and SurgBench~\cite{SurgBench} spans 22 procedures over 11 specialties. MedFrameQA~\cite{MedFrameQA} pioneers spatiotemporal modeling via cross-frame reasoning but lacks scale and fine-grained localization.
SurgCoT bridges this gap by unifying spatiotemporal reasoning, hierarchical knowledge, and localization supervision in a clinically validated framework, establishing a comprehensive evaluation baseline for surgical MLLMs under CoT reasoning.

%% file: sec/3_SurgBench5ST.tex
\begin{figure*}[!t]
\centering
\includegraphics[width=1\textwidth]{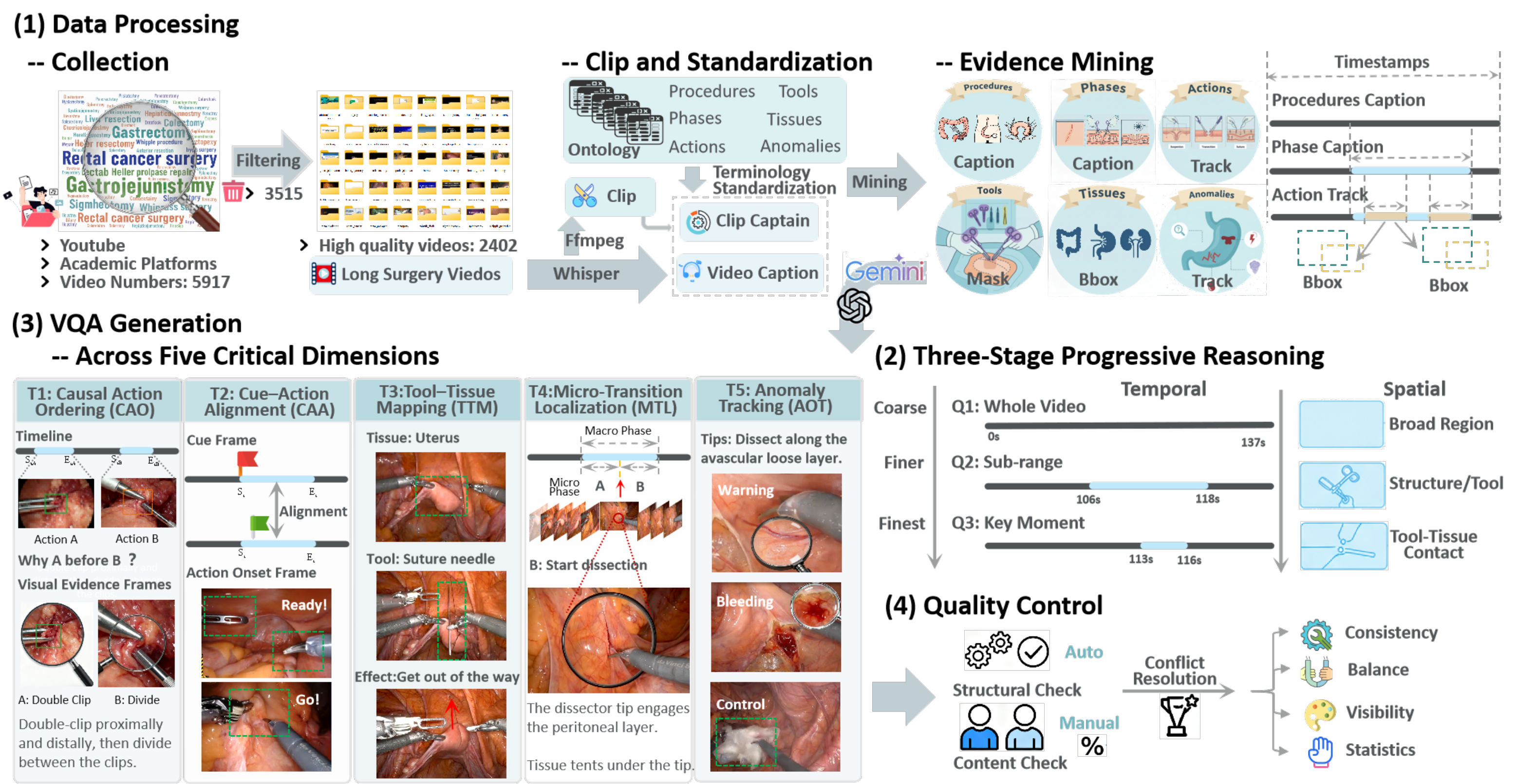}
\caption{Construction pipeline of SurgCoT benchmark.}
\label{pipeline}
\end{figure*}
\section{Construction of SurgCoT Benchmark} 
% \noindent\textbf{Overview.} \quad 
% SurgCoT is a fine-grained spatiotemporal evaluation framework for surgical MLLMs that spans \textbf{7} surgical specialties and \textbf{35} procedures, organized around five spatiotemporal task dimensions (5ST) to mirror expert reasoning along the surgical workflow.
% To probe region and time specific understanding, it adopts a three step (Q1$\rightarrow$Q2$\rightarrow$Q3) five tuple annotation protocol (\texttt{Question/Option/Knowledge/Clue/Answer}) in which the \emph{clue} progressively narrows the temporal and spatial scope, eliciting hierarchical reasoning and precise localization.
% This design enables rigorous measurement of MLLMs' fine-grained spatiotemporal integration, causality driven inference, and procedure aware comprehension across diverse real world surgical scenarios.

% \noindent\textbf{Overview.} \quad 
Fig.~\ref{pipeline} outlines the benchmark construction pipeline with expert oversight at each stage: 1)~\textbf{Data Processing} with multi-source video curation, hierarchical segmentation, and evidence mining; 2) \textbf{Three-Stage Progressive Reasoning} with five-tuple annotation protocol; 
~3)~\textbf{VQA Generation} producing 78,522 QA pairs via structured task templates and ontology-driven design; 4)~\textbf{Quality Control} via dual-pass validation and multi-criteria verification.

% \noindent\textbf{Overview.} \quad 
% Fig.~\ref{pipeline} illustrates the pipeline for benchmark construction. It involves three key steps, with expert in the loop at each step. 
% 1)~\textbf{Data Processing}, which involves comprehensive multi-source video collection and stringent quality filtering, followed by hierarchical clip segmentation with multimodal cue fusion, ASR-based temporal alignment, and ontology-driven terminology normalization, culminating in an end-to-end evidence mining framework that extracts spatiotemporally aligned annotations for surgical reasoning tasks.
% 2)~\textbf{VQA Generation}. In collaboration with clinical experts, we generate the SurgCoT dataset through a structured VQA generation framework comprising five spatiotemporal task families, each implemented via dedicated question templates under a three-step progressive protocol with ontology-driven distractor design, producing 446,535 multiple-choice QA pairs for surgical reasoning evaluation.
% 3)~\textbf{Quality Control}. Our quality control framework employs a dual-pass human-in-the-loop protocol combining automated structural validation with expert content review, targeting four key aspects—schema/logic consistency, category balance, visibility tagging, and statistical reliability—ensuring dataset integrity through majority-vote adjudication and stratified sampling.

\subsection{Data Processing}
\noindent\textbf{Data Collection.} \quad 
Our benchmark is constructed from a multi-source surgical video collection systematically organized across seven clinical specialties: Colorectal, Urological, Upper Gastrointestinal (Upper GI), Ocular, Gynecologic, General Surgery (GenSurgery) and Hepatobiliary-Pancreatic (HPB). 
The dataset integrates public platform content (\eg, YouTube, ASVIDE), ten open-source repositories, and proprietary clinical archives, yielding \textbf{8,917} initial surgical cases. Following standardized filtration prioritizing procedural completeness, clinical validity, and bilingual narration for temporal alignment, \textbf{2,841} high-quality videos (31.9\% of original corpus) were retained. All data were rigorously de-identified in compliance with ethical standards, with extended analyses available in Appendix.

\noindent\textbf{Clip and Standardization.} \quad 
Surgical videos are processed into standardized clips to enable fine-grained spatiotemporal reasoning: 1) \textbf{Multi-source segmentation} uses hierarchical cue fusion (visual scenes, tool/tissue transitions, and ASR anchors) to create semantically coherent segments; 2) \textbf{ASR alignment} produces millisecond-precise, temporally consistent captions without altering medical content; 3) \textbf{Ontology-driven normalization} maps surface terms to canonical entities using a controlled vocabulary, ensuring terminological consistency while preserving original captions. This workflow guarantees temporally aligned and semantically standardized clips.

\noindent\textbf{Evidence Mining.} \quad 
We retain public dataset annotations and introduce an end-to-end evidence mining paradigm to convert unannotated clips into spatiotemporally-grounded supervision units. 
1)~The pipeline utilizes ASR captions as semantic anchors, performs ontology-driven terminology normalization to create lexical-canonical mappings, and annotates \texttt{Procedures} and \texttt{Phases} by merging semantically consistent sentence windows. 
2)~\textbf{Spatial evidence} is constructed through frame-wise \texttt{Tissue} detection by YOLOv10~\cite{wang2024yolov10} and \texttt{Tool} segmentation by SAM2~\cite{ravi2024sam}, with cross-frame tracking via ByteTrack~\cite{zhang2022bytetrack}.  
3)~\textbf{Temporal evidence} is extracted by detecting \texttt{Action} onsets from appearance-change indicators, using minimal visual cues as frame-level anchors. \texttt{Anomalies} are annotated with tracks specifying onset time and minimal ROIs. 
All evidence is bidirectionally aligned with ASR timestamps, with caption windows split at event boundaries while preserving original timing. 
The generated annotations integrate \texttt{Procedure}/\texttt{Phase} captions, \texttt{Tool} masks, \texttt{Tissue} bounding boxes, and \texttt{Action}/\texttt{Anomaly} tracks, enabling progressive spatiotemporal reasoning through three-stage window narrowing and region-to-interface grounding for comprehensive VQA evaluation.

\subsection{Three-Stage Progressive Reasoning with Five-Tuple Annotation Protocol}
\label{sec:progressive_reasoning}
% \rjf{largely extend the discussion here. Explain how SurgCoT achieves CoT reasoning.}

% \red{To enable chain-of-thought reasoning in surgical video understanding, as shown in Fig.~\ref{overview}, we introduce a three-stage progressive reasoning framework (\texttt{Q1→Q2→Q3}) with a comprehensive five-tuple annotation protocol (\texttt{Question→Option→Knowledge→Clue→Answer}). The novelty is two-fold: 1) It establishes a hierarchical reasoning pathway progressing from video-level comprehension through clip-level analysis to frame-level localization. As shown in Fig.~\ref{overview}, all five annotation tuples evolve through these stages, with each stage's \texttt{Answer} serving as contextual evidence for the subsequent stage, forming a continuous reasoning chain. 
% 2) It employs synergistic annotation strategies where the \texttt{Knowledge} field provides clinically relevant background context based on the \texttt{Question} and \texttt{Option}, while the \texttt{Clue} supplies targeted spatiotemporal evidence extracted directly from the surgical video content. 
% This integrated design progressively narrows both temporal and spatial scope, effectively guiding MLLMs through structured hierarchical reasoning.} 

To enable structured, verifiable CoT reasoning in surgical video understanding, we introduce a three-stage progressive reasoning framework (\texttt{Q1→Q2→Q3}) with a comprehensive five-tuple annotation protocol (\texttt{Question→Option→Knowledge→Clue→Answer}). 

\noindent \textbf{Five-Tuple Annotation Protocol.} \quad
For each stage, we annotate five fields to collectively enable explicit, verifiable reasoning. 
\texttt{Question}: A clinically meaningful query aligned with surgical workflow;
\texttt{Option}: Curated, mutually exclusive candidates designed to disambiguate similar phenomena (\eg, tool reflections vs. true bleeding) and constrain the hypothesis space;
\texttt{Knowledge}: Provides domain priors (\eg, color/flow patterns, typical anatomy, device behavior) that explain clinical plausibility independent of video content;
\texttt{Clue}: Supplies video-grounded evidence (\eg, temporal windows, spatial ROIs, landmarks) that directs attention to case-specific diagnostic anchors;
\texttt{Answer}: the adjudicated target.  
Critically, \texttt{Answer} is carried forward as conditioning context for the next stage, enforcing causal dependency and spatiotemporal narrowing. 
Serializing \texttt{Knowledge} and \texttt{Clue} before \texttt{Answer} encourages a ``\textit{reason-then-decide}" pattern: \texttt{Knowledge} supplies clinical ``\textit{why}," while \texttt{Clue} anchors ``\textit{where/when}," yielding a transparent, video-grounded chain of thought. 

% Each stage is annotated with five fields that together induce explicit, verifiable reasoning. \textbf{Question} poses a clinically meaningful query aligned with the surgical workflow, belongs to one of five critical dimensions. \textbf{Option} provides curated, mutually exclusive candidates designed to disambiguate look alike phenomena (\eg, tool reflections vs.\ true bleeding) and to constrain the hypothesis space. \textbf{Knowledge} contributes \emph{domain priors}, such as color/flow patterns, typical anatomy/planes, and device behavior, that are \underline{clinical} rather than video specific, explaining \emph{why} particular choices are plausible. \textbf{Clue} supplies \emph{video grounded evidence}, namely a tight temporal window and spatial region derived from the case (timestamps, ROI labels, landmarks), offering the \underline{case specific} pointer that directs model attention to the relevant evidence. Finally, \textbf{Answer} is the adjudicated target for the stage; critically, $\texttt{Answer}_{k}$ is carried forward as conditioning context for $\texttt{Q}_{k+1}$, enforcing causal dependency and monotonic narrowing of time and space. In practice, serializing \texttt{Knowledge} and \texttt{Clue} before \texttt{Answer} encourages a "reason-then-decide" pattern, where \emph{Knowledge} supplies the clinical "why" and \emph{Clue} anchors the "where/when," yielding a transparent chain of thought grounded in the surgical video. 

\noindent \textbf{Three Stages of Progressive Reasoning.} \quad
The framework operationalizes clinical reasoning by decomposing complex spatiotemporal diagnostics into three hierarchically interdependent stages, progressively focusing subproblems while conditioning each stage on validated evidence from its predecessor. 
\textbf{Q1.~Video-Level Comprehension}: identifies high-level clinical events (\eg, ``\textit{active bleeding present?}") across the surgical video, establishing a global hypothesis space by determining whether a target phenomenon occurs. Its output constrains the reasoning trajectory and provides the primary conditioning signal for subsequent stages. 
\textbf{Q2.~Video-Level Comprehension}: operates under Q1's validated output, performing spatiotemporal localization by determining when a target event (\eg, \textit{active bleeding}) first emerges within a bounded temporal window and where it occurs at region-of-interest granularity. This process prunes the hypothesis space from video-level to a focused spatiotemporal segment, providing temporal anchors and spatial regions-of-interest for the next stage. 
\textbf{Q3.~Frame-/Patch-Level Localization}: strictly conditioned on Q2's spatiotemporal bounds, requiring precise, fine-grained localization of exact onset frames and anatomical sites (\eg, suture hole vs. adjacent tissue) with pixel-/bbox-level precision. Its output enables detailed auditability and provides supervision for downstream tasks. 
The cascading framework ensures each reasoning step is grounded in validated prior evidence, enforcing logical coherence and traceability throughout the diagnostic workflow.

\noindent\textbf{Progressive Conditioning Reasoning Chain.} \quad
We establish a rigorous context-carry mechanism, which enforces explicit logical dependencies across three refined reasoning stages, with each stage building strictly upon its predecessor's validated constraints:  
% We implement a rigorous context-carry mechanism that establishes explicit logical dependencies across three progressively refined reasoning stages, ensuring each subsequent stage operates strictly within constraints validated by its predecessor: 
\begin{equation*}
\begin{aligned}
&\underbrace{(\texttt{Q1},\ \texttt{O1},\ \texttt{K1},\ \texttt{C1})}_{\text{Global Video Comprehension}}
 \Rightarrow \texttt{A1} \\
& \quad \to\ \underbrace{(\texttt{Q2},\ \texttt{O2},\ \texttt{K2},\ \texttt{C2},\ \texttt{A1})}_{\text{Conditioned Clip Analysis}}
 \Rightarrow \texttt{A2} \\
& \quad \quad \to\ \underbrace{(\texttt{Q3},\ \texttt{O3},\ \texttt{K3},\ \texttt{C3},\ \texttt{A2})}_{\text{Fine-grained Frame Localization}} 
 \Rightarrow \texttt{A3}.
\end{aligned}
\end{equation*}
The chain operates through three critical dependency mechanisms: 
\textbf{1)~Semantic Constraint Propagation}: Each stage's output forms an immutable foundation for the next. For instance, if \texttt{Q1} confirms ``\textit{active bleeding present}," \texttt{Q2} must proceed within this context without reconsideration, ensuring logical consistency and narrative coherence.
\textbf{2)~Spatiotemporal Scope Refinement}: The analysis scope narrows progressively. \texttt{Q1} processes the full video (minutes), \texttt{Q2} focuses on relevant clips (seconds-minutes), and \texttt{Q3} zooms into critical frames (sub-second) and anatomical regions (pixel/bbox level), mirroring the clinical workflow of identifying and then localizing abnormalities. 
\textbf{3) Evidence Accumulation and Validation}: \texttt{Knowledge} evolves from general anatomy (\texttt{K1}) to lesion-specific details (\texttt{K3}), while clues progress from temporal landmarks (\texttt{C1}) to spatial regions (\texttt{C2}) and pixel-level evidence (\texttt{C3}), reflecting increasing diagnostic certainty. 
This design enforces a strict chain of dependencies for diagnostic decisions. For instance, a surgical recommendation (\texttt{A3}) must be grounded in a confirmed lesion location (\texttt{A2}), which itself requires established pathology (\texttt{A1}). This creates an auditable trail that mirrors clinical reasoning, ensuring treatment is justified by diagnostic evidence. The process guarantees progressively decreasing uncertainty and evaluates both diagnostic accuracy and adherence to the required logical progression.

% This design ensures that diagnostic choices are logically invalid without supporting evidence from prior stages. For example, recommending surgical intervention (\texttt{A3}) requires confirmed lesion location (\texttt{A2}), which itself depends on established pathology presence (\texttt{A1}). This creates an auditable trail mirroring clinical standards where treatment must be justified by diagnostic findings. 
% The monotonic narrowing property guarantees decreasing uncertainty at each stage, while embedded dependencies prevent diagnosis guessing without demonstrated intermediate reasoning. The framework evaluates both diagnostic accuracy and adherence to the structured reasoning process expected of clinical experts. 

% \noindent \textbf{Progressive Conditioning Reasoning Chain.} \quad
% We implement a strict \emph{context carry} mechanism:
% \begin{equation*}
% \begin{aligned}
% &\underbrace{(\texttt{Q1},\ \texttt{O1},\ \texttt{K1},\ \texttt{C1})}_{\text{global}}
%  \Rightarrow \texttt{A1} \\
% &\Rightarrow\ \underbrace{(\texttt{Q2},\ \texttt{O2},\ \texttt{K2},\ \texttt{C2},\ \texttt{A1})}_{\text{conditioned}}
%  \Rightarrow \texttt{A2} \\
% &\Rightarrow\ \underbrace{(\texttt{Q3},\ \texttt{O3},\ \texttt{K3},\ \texttt{C3},\ \texttt{A2}}_{\text{fine-grained}}
%  \Rightarrow \texttt{A3}.
% \end{aligned}
% \end{equation*}
% This design yields a monotonic narrowing of (time, space) and a \emph{dependency chain} in which later choices are invalid without earlier evidence and an explicit, auditable CoT. 

\subsection{VQA Generation: Five Critical Dimensions} 
With clinical experts, we define five spatiotemporal reasoning tasks to investigate deep cognitive processes. Each task uses the three-stage framework (Sec.~\ref{sec:progressive_reasoning}) to generate ontology-driven VQA pairs containing spatial, temporal, and semantic distractors. This yields the SurgCoT benchmark, comprising 19,345 main and 59,177 sub-questions, for evaluating hierarchical reasoning in surgical videos.

% In collaboration with clinical experts, we define five spatiotemporal reasoning tasks to probe deep cognitive processes beyond surface-level analysis. Each task is implemented via a three-stage progressive framework with five-tuple annotation (Sec.~\ref{sec:progressive_reasoning}), generating ontology-driven VQA pairs with spatial, temporal, and semantic distractors. The resulting dataset of 19,345 main questions and 59,177 sub-questions establishes the SurgCoT benchmark for evaluating hierarchical reasoning in surgical videos. 

\noindent\textbf{Causal Action Ordering (CAO).} \quad 
This task determines the causal sequence of surgical micro-actions by using minimal visual evidence and fine-grained spatiotemporal cues (\eg, action onset, interface-level ROIs) to enable verifiable rationale inference. Example: \textit{\textbf{Q}: In cholecystectomy, which occurs first: clip placement or scissor division? \textbf{A}: Clip placement precedes and enables subsequent division.}

% CAO determines and justifies the causal temporal order between surgical micro-actions using minimal visible evidence at the tool–tissue interface. Moving beyond mere ordering, it relies on fine-grained spatiotemporal supervision, including action onset timing, interface-level ROIs, and causal visual cues, to enable verifiable rationale inference. For example: \textit{\textbf{Q}: In cholecystectomy, which occurs first: clip placement or scissor division? \textbf{A}: Clip placement precedes and enables subsequent division.}

\noindent\textbf{Cue-Action Alignment (CAA).} \quad 
CAA aligns pre-action cues with micro-actions to pinpoint action initiation in time and space, thus disambiguating intent from execution and providing verifiable onset anchors, a critical capability for surgical video reasoning. Example: \textit{\textbf{Q}: When does visible execution first begin? \textbf{A}: Around $t=425$s, initiating at the hook–peritoneum interface over Calot’s triangle.}

% CAA aligns pre-action cues with subsequent micro-actions, determining when the action becomes visible and where it initiates at the tool–tissue interface. This capability is critical for surgical video reasoning as it disambiguates intent from execution and provides verifiable temporal anchors for action onset. For example: \textit{\textbf{Q}: When does visible execution first begin? \textbf{A}: Around $t=425$s, initiating at the hook–peritoneum interface over Calot’s triangle.}

\noindent\textbf{Affordance Mapping (AM).} \quad 
AM formalizes tool–tissue interactions by grounding them with spatiotemporal evidence and modeling their relationships, enabling progressive reasoning from coarse recognition to fine-grained localization. Example: \textit{\textbf{Q}: Identify the first affordance event that improves exposure. \textbf{A}: uterine suspension at the anterior serosal needle entry ($\sim$485s).}

% AM formalizes tool–tissue interactions by grounding them at the interface level with spatiotemporal evidence. By modeling tool, tissue, and effect relationships, it enables progressive reasoning from coarse identification to fine-grained localization, providing a reproducible basis for evaluation. For example:  \textit{\textbf{Q}: Identify the first affordance event that improves exposure. \textbf{A}: (needle driver + curved suture needle, uterus, displacement), initiating at the needle–uterine serosa interface on the anterior wall ($\approx$485s).}

\noindent\textbf{Micro-Transition Localization (MTL).} \quad 
MTL identifies frame-level boundaries between micro-phases using minimal visual evidence, providing verifiable spatiotemporal evidence of transitions to enable fine-grained temporal reasoning. Example: \textit{\textbf{Q}: At what moment does the micro-phase shift from positioning to dissection? \textbf{A}: At $\approx$312s, when the hook–peritoneum interface over Calot’s triangle first incises and tents the peritoneal layer.}

% MTL identifies the precise frame-level boundary between adjacent micro-phases within a surgical macro-phase, using minimal visual evidence at the tool–tissue interface. It provides verifiable spatiotemporal evidence that specifies both the timing of the transition and the precise anatomical location where the decisive cue occurs, enabling fine-grained temporal reasoning in surgical workflows. For example: \textit{\textbf{Q}: At what exact moment does the micro-phase shift from positioning to dissection? \textbf{A}: At $\approx$312s at the hook–peritoneum interface over Calot’s triangle, when the dissector tip engages the peritoneal layer and causes tissue tenting.}

\noindent\textbf{Anomaly Onset Tracking (AOT).} \quad 
AOT identifies the onset and early trajectory of surgical anomalies, providing precise spatiotemporal localization to enable evidence-based detection and assessment of workflow deviations.  Example: \textit{\textbf{Q}: Can you precisely localize the anomaly onset in this clip? \textbf{A}: At 647.0s, Spatial: [540, 420, 80, 60], where bleeding first appears at the cystic artery stump before tracking its spread until hemostasis.}

% AOT identifies the first visible onset of a surgical anomaly in both time and space, and optionally tracks its early spread and control evidence. It provides precise spatiotemporal localization (onset time, ROI, and trajectory) for anomalies, enabling evidence-based detection and assessment of deviations from the normal surgical workflow. For example: \textit{\textbf{Q}: Can you precisely localize the anomaly onset in this clip? \textbf{A}: At 647.0s, Spatial: [540, 420, 80, 60], where bleeding first appears at the cystic artery stump before tracking its spread until hemostasis.}

These five dimensions form a cohesive framework for surgical reasoning assessment: \textbf{CAO} establishes causal logic; \textbf{CAA} anchors intent to execution; \textbf{AM} grounds interactions physically; \textbf{MLT} tracks micro-scale progressions; and \textbf{AOT} monitors pathway deviations. Together, they enable holistic evaluation from normative workflow reasoning (\textbf{CAO}, \textbf{CAA}, \textbf{AM}, \textbf{MTL}) to abnormal scenario handling (\textbf{AOT}) within a unified paradigm.

\subsection{Quality Control Protocol} 
The protocol integrates \textit{expert validation} with \textit{multi-criteria verification} via a dual-pass human-in-the-loop process. Automated structural validation ensures schema integrity and logical consistency (\eg, temporal/spatial progression, no answer leakage), while dual-annotator content validation with expert adjudication verifies task dependency, rationale verifiability, and ontology compliance. The protocol addresses four quality dimensions: \textit{Consistency} (schema/logic checks), \textit{Balance} (class-wise quotas), \textit{Visibility} (obstruction tagging for difficulty filtering), and \textit{Statistical Reliability} (inter-rater agreement metrics and acceptance thresholds). Disagreements are resolved by majority voting, with inconsistent samples discarded to ensure robust annotation.

\begin{figure}[!t]
\centering
\includegraphics[width=1\linewidth]{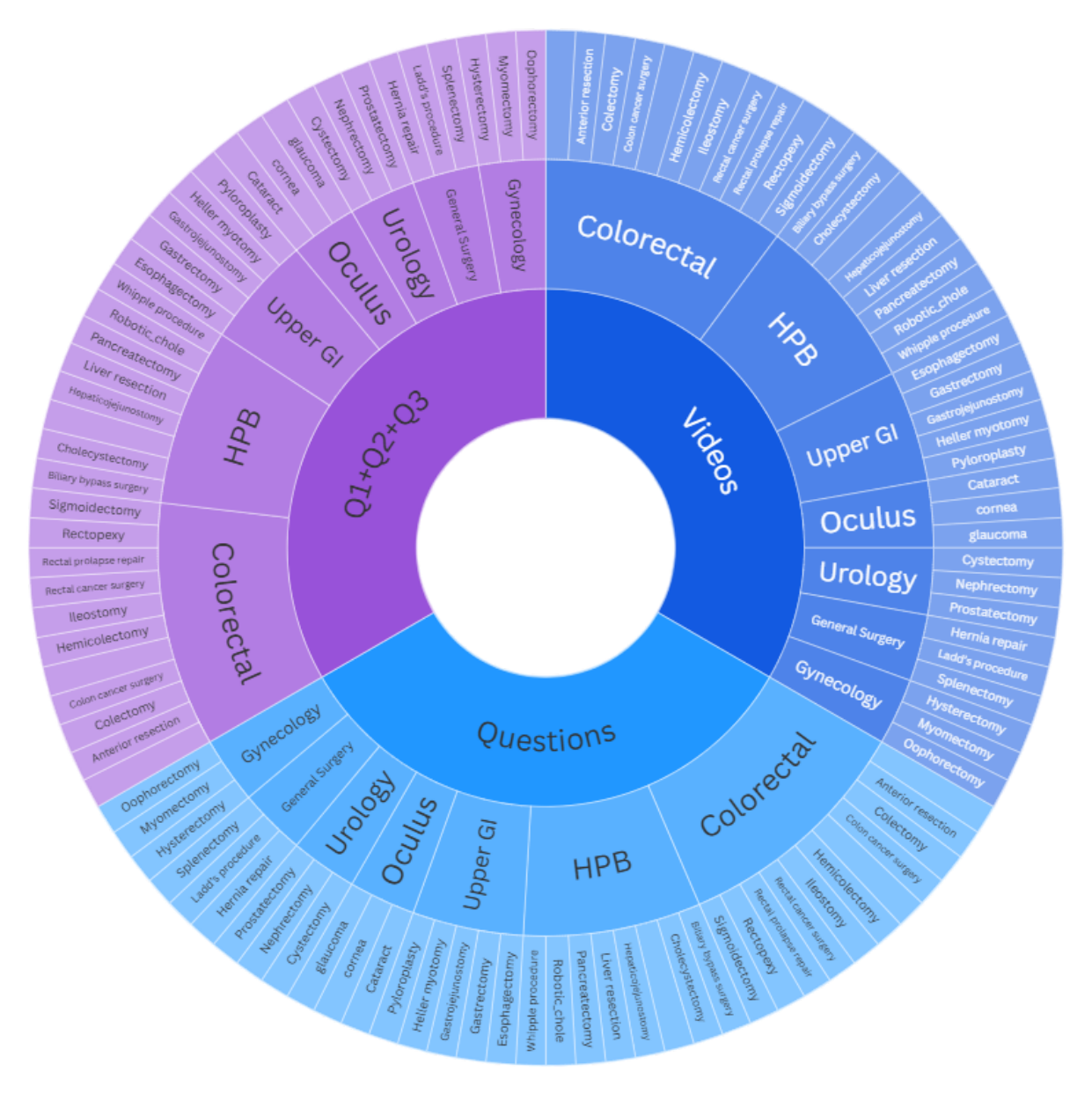}
\caption{Statistics of SurgCoT: 2,841 videos, 19,345 questions, and 59,177 sub-questions across 35 procedures and 7 specialties. 
}
\label{Sta}
\end{figure}
\subsection{Dataset Statistics}
SurgCoT contains 2,841 surgical videos across 35 procedures in 7 specialties (Fig.~\ref{Sta}), covering abdominal, pelvic, and ophthalmic surgeries. The benchmark includes 19,345 main questions and 59,177 sub-questions, averaging 7 main and 21 sub-questions per video. Question distribution aligns with specialty coverage, ensuring balanced supervision density while maintaining a long-tail of rare procedures to evaluate both common and uncommon workflows.

% SurgCoT comprises 2,841 surgical videos spanning 35 procedures across 7 major specialties, covering a diverse spectrum of abdominal, pelvic, and ophthalmic surgeries as shown in Fig.~\ref{Sta}. The benchmark includes 19,345 main questions and 59,177 sub-questions (Q1+Q2+Q3), averaging 7 main and 21 sub-questions per video. Question distribution aligns with video coverage across specialties, ensuring balanced supervision density while maintaining a long-tail of less common procedures to support evaluation of both routine and rare surgical workflows. 

% \subsection{Statistics of SurgCoT}
% SurgCoT comprises \textbf{2{,}841} surgical videos, covering 36 procedures from 7 major categories (Colorectal, HPB, Oculus, Upper GI, Gynecology, General Surgery, and Urology). As shown in Fig.~\ref{Sta}, the dataset spans a broad range of abdominal, pelvic, and ophthalmic procedures, offering comprehensive coverage of both high-volume routine operations and more specialized oncologic and reconstructive surgeries.
% In total, the benchmark includes \textbf{22{,}801} main questions and \textbf{69{,}729} sub-questions (Q1+Q2+Q3), i.e., on average about 8 main questions and 25 sub-questions per video. The distribution of questions closely follows that of videos across categories, ensuring comparable supervision density for different specialties. Within each category, common procedures contribute to most samples, while the remaining procedures form a long tail, making SurgCoT suitable for evaluating both common and relatively rare surgical workflows.

%% file: sec/4_Experiments.tex
\begin{table*}[t]
\centering
\caption{
Evaluation of 10 MLLMs across five clinical reasoning tasks under progressive settings (\textbf{BL}$\rightarrow$\textbf{KE}$\rightarrow$\textbf{FC}) shows: 1) commercial models outperform open-source and medical-specialized counterparts, and 2) the five-tuple annotation protocol improves reasoning accuracy under both (\textbf{KE}) and (\textbf{FC}) settings.
Best results in \textbf{bold}, second-best \underline{underlined}.
}
\label{tab:main_results}
\resizebox{\linewidth}{!}{
\begin{tabular}{clccccccccccccccccccc} % l + 14 c = 15 columns total
\toprule
& \multirow{2}{*}{\textbf{Model}} &
\multicolumn{3}{c}{\textbf{CAO. (\%) $\uparrow$}} &\multicolumn{3}{c}{\textbf{CAA. (\%) $\uparrow$}} &\multicolumn{3}{c}{\textbf{AM. (\%) $\uparrow$}} &\multicolumn{3}{c}{\textbf{MTL.(\%) $\uparrow$}} &\multicolumn{3}{c}{\textbf{AOT. (\%) $\uparrow$}} &\multicolumn{3}{c}{\textbf{Avg. (\%) $\uparrow$}} \\
\cmidrule(lr){3-5} \cmidrule(lr){6-8} \cmidrule(lr){9-11} \cmidrule(lr){12-14} \cmidrule(lr){15-17} \cmidrule(lr){18-20}
& & \textbf{BL} & \textbf{KE} & \textbf{FC} & \textbf{BL} & \textbf{KE} & \textbf{FC} & \textbf{BL}
& \textbf{KE} & \textbf{FC} & \textbf{BL} & \textbf{KE}
& \textbf{FC} & \textbf{BL} & \textbf{KE} & \textbf{FC} & \textbf{BL} & \textbf{KE} & \textbf{FC} \\
\midrule
\multirow{3}{*}{\rotatebox[origin=c]{90}{\textbf{\textit{Com.}}}} & 
GPT-5~\cite{openai2025gpt5}  
& 85.07 & 91.56 & 94.79 & \underline{80.99} & 85.29 & \underline{90.64} & \textbf{78.97} & \textbf{77.67} & 82.34 & 69.14 & 81.63 & 86.71 & \underline{68.95} & 66.53 & 78.23 & \textbf{76.62} & 80.54 & \textbf{87.58} \\
& Gemini-2.5-Pro~\cite{google2024gemini} 
& 89.66 & 93.99 & 96.92 & 77.97 & 76.75 & 84.62 & 60.31 & 73.20 & \underline{86.55} & 69.82 & \underline{82.73} & 84.05 & 52.32 & \underline{76.48} & \underline{80.85} & 70.02 & \textbf{81.83} & 87.20 \\
& claude-sonnect-4.5~\cite{claude_sonnet_4_5} 
& \textbf{92.81} & \textbf{97.99} & \underline{98.33} & \textbf{83.66} & \underline{86.46} & 89.12 & 59.80 & 68.10 & \textbf{87.51} & \textbf{79.49} & 72.45 & 86.39 & 54.74 & 67.35 & 75.34 & 74.10 & 78.87 & \underline{87.54} \\
\midrule
\multirow{4}{*}{\rotatebox[origin=c]{90}{\textbf{\textit{Medical}}}} & 
MedGemma-27B-IT~\cite{sellergren2025medgemma}   
& 83.67 & 94.99 & 96.99 & 74.02 & 81.28 & 86.49 & \underline{73.39} & 70.83 & 79.26 & 72.69 & 75.54 & 85.60 & 51.02 & 59.27 & 80.52 & 70.96 & 76.37 & 86.37 \\
& Lingshu-7B~\cite{xu2025lingshu} 
& 86.68 & \underline{97.24} & 97.50 & 65.83 & 64.57 & 72.27 & 49.06 & 53.98 & 63.50 & 65.12 & 68.99 & 73.84 & 50.46 & 62.46 & 67.64 & 63.43 & 69.45 & 74.95 \\ 
& LLaVA-Med-7B~\cite{li2023llava}  
& 89.95 & 96.93 & 97.50 & 66.10 & 72.34 & 71.52 & 62.29 & \underline{74.68} & 76.16 & 71.43 & 81.84 & \textbf{88.67} & 50.97 & 47.25 & 74.81 & 68.15 & 75.22 & 81.73 \\
& HuatuoGPT-Vision-7B~\cite{chen2024huatuogpt} 
& 78.47 & 80.53 & 88.74 & 68.28 & 65.91 & 73.24 & 67.24 & 68.63 & 81.37 & 72.53 & 72.69 & \underline{87.37} & 63.89 & 68.21 & 77.95 & 70.08 & 71.19 & 83.73 \\ 
\midrule
\multirow{3}{*}{\rotatebox[origin=c]{90}{\textbf{\textit{Open.}}}}
& InternVL-8B~\cite{chen2024internvl}   
& \underline{90.72} & 94.17 & 97.53 & 70.08 & 73.88 & 82.75 & 59.12 & 59.45 & 68.46 & 68.35 & 69.52 & 84.85 & 51.46 & 65.90 & 76.03 & 67.95 & 73.58 & 82.32 \\
& Qwen2.5-VL-7B~\cite{wang2024qwen2}   
& 84.42 & 87.17 & \textbf{98.98} & 65.64 & 68.77 & 83.56 & 60.95 & 66.78 & 68.41 & \underline{76.59} & \textbf{83.40} & 85.02 & 56.64 & 49.97 & 61.30 & 68.85 & 71.22 & 79.45 \\
& Qwen3-VL-8B~\cite{xu2025qwen3} 
& 87.96 & 92.69 & 96.77 & 75.51 & \textbf{87.42} & \textbf{91.57} & 60.67 & 66.75 & 73.30 & 74.07 & 76.10 & 83.30 & \textbf{79.00} & \textbf{77.46} & \textbf{86.68} & \underline{75.44} & \underline{81.48} & 86.92 \\
\bottomrule
\end{tabular}
}
\end{table*}

\section{Experimental Results}

\subsection{Experimental Setup}
\noindent \textbf{Evaluated Models.} \quad 
We evaluate 10 MLLMs: \textbf{Commercial models} include GPT-5~\cite{openai2025gpt5}, Claude-Sonnet-4.5~\cite{claude_sonnet_4_5}, and Gemini-2.5-Pro~\cite{google2024gemini}, serving as leading multimodal baselines with strong performance on diverse vision-language tasks;  
\textbf{Open-source models} include Qwen2.5-VL-7B, Qwen3-VL-8B~\cite{xu2025qwen3} and Intern3.5-VL-8B~\cite{wang2025internvl3};
% , LLaVA-v1.5-13B~\cite{liu2023llava15}, and GLM-4.5V~\cite{hong2025glm};
%LLaMA3-LLaVA-Next-8B~\cite{llava_next_interleave_2024}; 
\textbf{Med-specialized  models} include MedGemma-27B-IT~\cite{sellergren2025medgemma}, Lingshu-7B~\cite{xu2025lingshu}, LLaVA-Med-7B~\cite{li2023llava}, 
and HuatuoGPT-Vision-7B~\cite{chen2024huatuogpt}. 
We use accuracy as the primary evaluation metric.

% \noindent\textbf{Human Study.}\quad 

% \noindent\textbf{Evaluation Metrics.}\quad 
% We use accuracy as our primary metric, calculated as the percentage of correct answers across test samples. %For \textbf{ Knowledge-Enhanced} and \textbf{Full-Context} settings, we report separate accuracies for the main question (final integrated outcome) and sub-questions (Q1–Q3; intermediate reasoning steps).  
% We adopt accuracy as our primary evaluation metric, computed as the percentage of correctly answered questions across all test samples. For the Knowledge-Enhanced and Full-Context settings, we report separate accuracies for the main question and the sub-question. Main Question: The final reasoning outcome requires integration of all preceding stages. Sub-Questions (Q1, Q2, Q3): Intermediate reasoning steps that form the reasoning chain.

\noindent \textbf{Evaluation Protocol.} \quad 
We design a three-stage protocol to systematically assess MLLMs’ CoT reasoning capabilities: 
\textbf{1) Baseline: (BL)}: Models receive only the surgical video and main question, evaluating end-to-end reasoning without guidance.
\textbf{2) Knowledge-Enhanced (KE)}: Adds clinical background knowledge; models answer both main and sub-questions (\texttt{Q1}, \texttt{Q2}, \texttt{Q3}), evaluate if \texttt{Knowledge} aids progressive reasoning.
\textbf{3) Full-Context (FC)}: Provides full annotation support (Video, \texttt{Knowledge} and \texttt{Clue}) evaluating the comprehensive diagnostic capability with complete reasoning scaffolds. 

\noindent \textbf{Implementation Details.} \quad 
We adopt unified prompts and decoding settings: a zero-shot template with fixed decoding parameters, including \(\mathrm{temperature}=0.0\), \(\mathrm{top\_p}=1.0\), \(\mathrm{max\_new\_tokens}=4096\), and \(\mathrm{repetition\_penalty}=1.0\); sampling retries are disabled. API-hosted models are invoked with these parameters, while local open-source models are run with \texttt{Torch 2.9.0} and \texttt{Transformers 4.57.1} on \texttt{CUDA 12.4}, using \texttt{bf16} inference on \(\text{NVIDIA A100 80\,GB} \times 8\) GPUs.

% Sampling retries are disabled. API-hosted models are invoked with these parameters, while local open-source models are run with  
% %\textit{Torch 2.8.0}, \textit{Transformers 4.57.1}, and \textit{vLLM 0.11.0} on \textit{CUDA 12.1}, using 
% batch size \(8\) and \textit{bf16} inference on 8 NVIDIA A100 80GB GPUs (single-GPU for \(\leq \)13B, multi-GPU for \(\geq \)30B models).

% We adopt unified prompts and decoding settings: a zero-shot template with fixed decoding parameters, including \(\mathrm{temperature}=0.0\), \(\mathrm{top\_p}=1.0\), \(\mathrm{max\_new\_tokens}=8192\) (increased to \(16384\) for cross-modal integration tasks), and \(\mathrm{repetition\_penalty}=1.0\); sampling retries are disabled. API-hosted models are invoked with these parameters, while local open-source models are run with \texttt{Torch 2.8.0}, \texttt{Transformers 4.57.1}, and \texttt{vLLM 0.11.0} on \texttt{CUDA 12.1}, using batch size \(8\) and \texttt{bf16} inference on \(\text{NVIDIA A100 80\,GB} \times 8\) GPUs (single-GPU for models \(\leq \)13B; multi-GPU for \(\geq \)30B).

\subsection{Evaluation of MLLMs}

Table~\ref{tab:main_results} summarizes the evaluation of 10 leading MLLMs, revealing four key observations. 

\noindent \textbf{Commercial Models Lead Peers.} \quad
As demonstrated in Table~\ref{tab:main_results}, commercial models consistently achieve superior performance compared to both open-source and medically-specialized counterparts across all five clinical reasoning dimensions. Leading commercial systems including GPT-5~\cite{openai2025gpt5}, claude-sonnect-4.5~\cite{claude_sonnet_4_5}, and Gemini-2.5-Pro~\cite{google2024gemini} demonstrate particularly strong capabilities in progressive reasoning scenarios (\textbf{BL} → \textbf{KE} → \textbf{FC}), %leveraging their advanced natural language understanding to effectively process and integrate complex multi-step reasoning chains, 
\ie, there is a stable performance progression as additional knowledge elements (\textbf{KE}) and spatiotemporal constraints (\textbf{FC}) are introduced. 
GPT-5~\cite{openai2025gpt5}, for instance, shows consistent performance gains across all task categories (on average +10.96\%), while Claude-Sonnet-4.5~\cite{claude_sonnet_4_5} demonstrates notable improvements in fine-grained decision-making tasks such as AOT (+20.60\%) and AM (+27.71\%). Conversely, open-source and medically specialized models display significant limitations in tasks requiring precise temporal and spatial reasoning, particularly struggling with the integration of multi-modal information sources.

% As demonstrated in Table~\ref{tab:main_results}, commercial models consistently outperform their open-source and medical counterparts across all five clinical reasoning dimensions. Models like GPT-5, Gemini-2.5-Pro, and Claude-Sonnet-4.5~\cite{claude_sonnet_4_5} exhibit strong performance across progressively reasoning settings (BL → KE → FC). These models leverage robust language reasoning capabilities, allowing them to process and integrate complex reasoning chains with greater ease than their open-source and medical counterparts.
% The commercial models also demonstrate a more stable performance improvement as additional knowledge (KE) and spatiotemporal cues (FC) are introduced, highlighting their superior ability to handle multi-dimensional reasoning tasks. For instance, GPT-5~\cite{openai2025gpt5}shows a relatively smooth increase in performance across all tasks, while other commercial models like Claude-Sonnet-4.5~\cite{claude_sonnet_4_5} also demonstrate significant improvements, especially in tasks requiring fine-grained decision-making, such as AOT and AM.
% In contrast, open-source and medical models struggle with lower accuracy, particularly in tasks requiring detailed temporal and spatial reasoning, reflecting their limitations in integrating multi-source information effectively. The differences in performance between commercial models and their peers underline the enhanced language understanding and reasoning capabilities in commercial MLLMs, which are well-suited for complex, multi-step clinical reasoning tasks. 

\begin{figure*}[!t]
\centering
\includegraphics[width=1\textwidth]{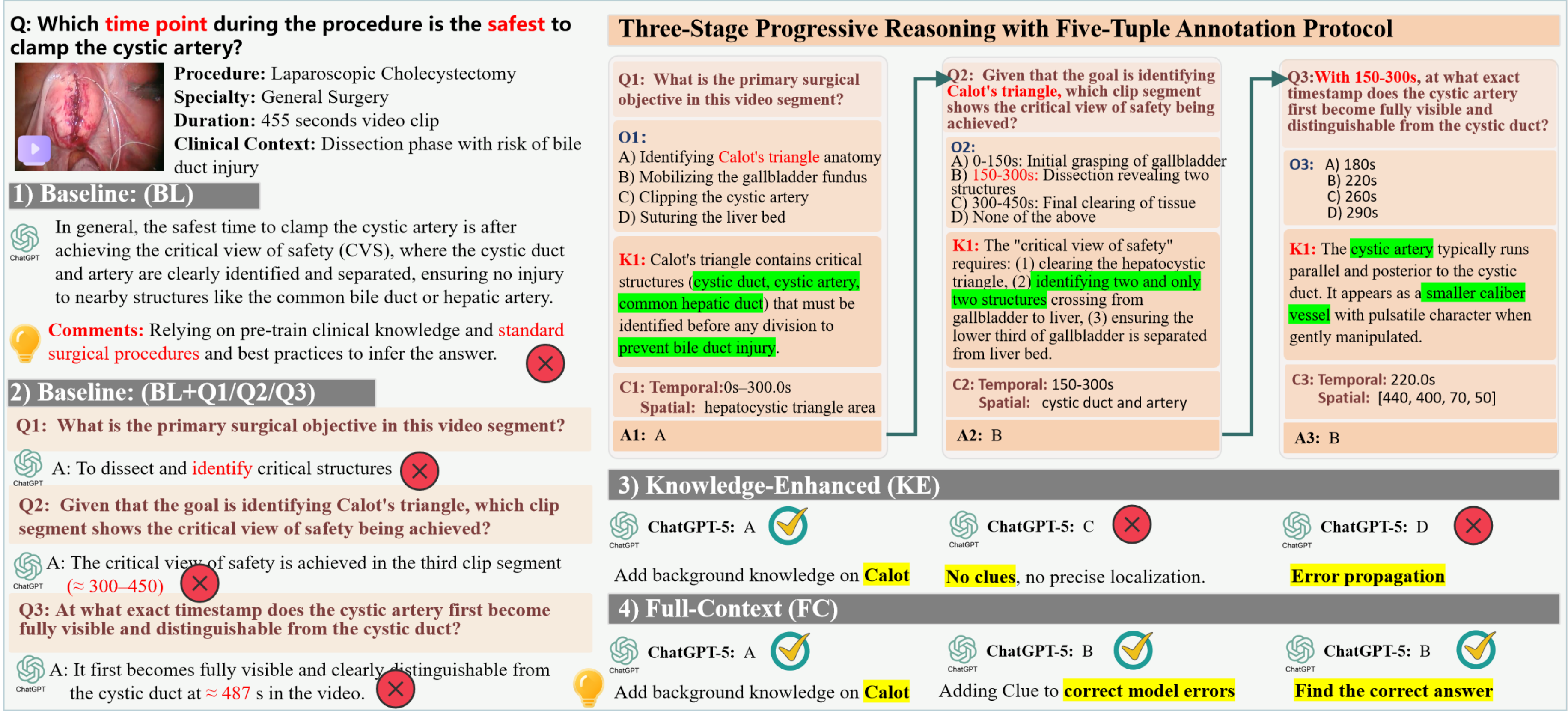}
\caption{
SurgCoT constructs a diagnostic chain-of-thought by progressively decomposing a flawed baseline (\textbf{BL}) into clinical sub-questions, correcting semantics with knowledge (\textbf{KE}), and refining evidence with spatiotemporal clues (\textbf{FC}). 
% SurgCoT's structured annotations (\texttt{Knowledge} and \texttt{Clue}) guide progressive reasoning: starting from a flawed baseline (\textbf{BL}), decomposing it into sub-questions to establish a clinically-grounded reasoning chain of thought, knowledge enhancement (\textbf{KE}) corrects understanding, and spatiotemporal clues (\textbf{FC}) refine it with visual evidence for accurate diagnostic.
}

%Representative case of SurgCoT illustrating how progressively adding \textbf{Knowledge} and \textbf{Clue} helps the model correct its initial mistaken reasoning based only on video+question, refine its intermediate steps, and ultimately produce a more accurate final answer.}
\label{case}
\end{figure*}

\noindent \textbf{Five-Tuple Annotation Protocol Boosts Accuracy.} \quad
Medical-specific models such as LLaVA-Med-7B~\cite{li2023llava} exhibit substantial performance gains when transitioning from \textbf{BL} to \textbf{KE}, with an average accuracy improvement of nearly 7\% across the five reasoning dimensions. This pronounced improvement indicates that explicit knowledge augmentation effectively compensates for domain-specific limitations. In contrast, commercial models like GPT-5~\cite{openai2025gpt5} show more moderate improvements (nearly 4\%), suggesting their stronger inherent language reasoning capabilities allow for more seamless knowledge integration.

The transition from \textbf{KE} to \textbf{FC} further demonstrates the critical importance of spatiotemporal grounding, %Models capable of effectively leveraging fine-grained spatiotemporal cues achieved significant additional gains. 
\eg, Qwen2.5-VL-7B~\cite{xu2025qwen3} showed a notable 8.23\% increase, highlighting its enhanced ability to utilize temporal and spatial clues for refined reasoning. Similarly, Claude-Sonnet-4.5~\cite{claude_sonnet_4_5}, despite strong baseline performance (BL = 74.10\%), improved by approximately 13.44\% with spatiotemporal clues, underscoring its advanced capacity for multimodal fusion. These observations collectively validate that the five-tuple annotation framework enables progressive spatiotemporal reasoning through structured cue enhancement across three reasoning settings: \textbf{BL}, \textbf{KE}, and \textbf{FC}. 

\begin{table}[t]
\centering
\footnotesize
\caption{Sub-question accuracy under progressive reasoning settings (\textbf{BL}$\rightarrow$\textbf{KE}$\rightarrow$\textbf{FC}). Evaluation reveals: 1) MLLMs show chain-of-thought reasoning gaps, with performance dropping at intermediate steps; 2) SurgCoT’s structured framework supports progressive reasoning, yielding gains under \textbf{KE}/\textbf{FC} settings.
Best results in \textbf{bold}, second-best \underline{underlined}.
}
\label{tab:Q_results}
\resizebox{\linewidth}{!}{
\begin{tabular}{clccccccccc}
\toprule
& \multirow{2}{*}{\textbf{Model}} &
\textbf{BL} & \multicolumn{4}{c}{\textbf{KE}} &\multicolumn{4}{c}{\textbf{FC}} \\
\cmidrule(lr){3-3} \cmidrule(lr){4-7} \cmidrule(lr){8-11}
& & \textbf{Q} & \textbf{Q} & \textbf{Q1} & \textbf{Q2} & \textbf{Q3} & \textbf{Q} & \textbf{Q1}
& \textbf{Q2} & \textbf{Q3}\\
\midrule
\multirow{3}{*}{\rotatebox[origin=c]{90}{\textbf{\textit{Com.}}}} 
& GPT-5~\cite{openai2025gpt5}
& \textbf{76.62} & 80.54 & 54.71 & 55.85 & 47.60 & \textbf{87.58} & 56.12 & 56.15 & 50.51 \\
& Gemini-2.5-Pro~\cite{google2024gemini} 
& 70.02 & \textbf{81.83} & 34.15 & 44.04 & 43.06 & 87.20 & 35.85 & 34.02 & 52.32 \\
& Claude-Sonnect-4.5~\cite{claude_sonnet_4_5} 
& 74.10 & 78.87 & 49.38 & 41.42 & 36.50 & \underline{87.54} & 53.04 & 34.85 & 30.44 \\
\midrule
\multirow{4}{*}{\rotatebox[origin=c]{90}{\textbf{\textit{Medical}}}} & 
MedGemma-27B-IT~\cite{sellergren2025medgemma}   
& 70.96 & 76.37 & 54.82 & 56.75 & \textbf{76.25} & 86.37 & 61.29 & \textbf{71.25} & \underline{70.42} \\
& Lingshu-7B~\cite{xu2025lingshu}  
& 63.43 & 69.45 & 36.06 & 47.93 & 43.53 & 74.95 & 48.36 & 58.58 & 56.52 \\
& LLaVA-Med-7B~\cite{li2023llava}  
& 68.15 & 75.22 & 31.71 & 36.65 & 36.46 & 81.73 & 34.40 & 39.75 & 37.57 \\
& HuatuoGPT-Vision-7B~\cite{chen2024huatuogpt} 
& 70.08 & 71.19 & \underline{69.97} & \textbf{67.69} & 55.62 & 83.73 & \underline{72.13} & \underline{70.31} & 69.00 \\
\midrule
\multirow{3}{*}{\rotatebox[origin=c]{90}{\textbf{\textit{Open}}}}
& Qwen3-VL-8B~\cite{xu2025qwen3} 
& \underline{75.44} & \underline{81.48} & \textbf{72.90} & 51.00 & \underline{73.04} & 86.92 & \textbf{74.75} & 67.49 & \textbf{70.98} \\
& Qwen2.5-VL-7B~\cite{wang2024qwen2} 
& 68.85 & 71.22 & 31.85 & 27.02 & 25.33 & 79.45 & 38.14 & 29.87 & 36.80 \\
& InternVL-8B~\cite{chen2024internvl} 
& 67.95 & 73.58 & 67.80 & \underline{65.79} & 48.86 & 82.32 & 64.89 & 55.28 & 51.63 \\
\bottomrule
\end{tabular}
}
\end{table}

\noindent \textbf{MLLMs Demonstrate Limitations in CoT Reasoning} \quad
To systematically evaluate CoT's reasoning capabilities, we analyze MLLM's performance across the three progressive reasoning stages (Q1, Q2, Q3) that form the foundational substructure of each main question (Q). As quantified in Table~\ref{tab:Q_results}, a pronounced performance dissociation emerges between main question accuracy and sub-question proficiency, revealing fundamental limitations in progressive reasoning. 

Notably, GPT-5~\cite{openai2025gpt5} achieves 76.62\% accuracy on main questions (Q) but exhibits a substantial performance degradation in sub-questions, dropping to 47.60\% in Q3. This pattern is consistently observed across other leading commercial models, including Gemini-2.5-Pro~\cite{google2024gemini} and Claude-Sonnet-4.5~\cite{claude_sonnet_4_5}. %, indicating a systemic weakness in maintaining reasoning consistency across multi-step inference processes. %The performance dissociation suggests that while these models can generate plausible final answers, they frequently lack the structured reasoning pathways necessary to support intermediate conclusions. 
Open-source models such as Qwen2.5-VL-7B~\cite{xu2025qwen3} demonstrate even more pronounced deficiencies, particularly in the initial (Q1) and final (Q3) sub-questions, underscoring their limited capacity for coherent multi-step inference. \textbf{\textit{These observations collectively highlight a critical gap in current MLLMs' ability to sustain logically consistent reasoning chains}}, a capability essential for reliable clinical decision-making where each diagnostic step must follow rigorously from preceding evidence.

% To analyze the reasoning ability in a chain-of-thought manner, we evaluate MLLMs for sub-question accuracy in each of the three progressive reasoning stages. 
% As shown in Table~\ref{tab:Q_results}, most MLLMs exhibit strong performance on main questions but struggle with sub-questions, highlighting limitations in their chain-of-thought reasoning abilities. For instance, GPT-5~\cite{openai2025gpt5} achieves a high accuracy of 73.29\% on the main question (Q), but its performance drops significantly in the sub-questions (Q1, Q2, Q3), with accuracy as low as 49.03\% in Q3. Similar trends are observed in other commercial models, such as Gemini-2.5-Pro and Claude-Sonnet-4.5, which also demonstrate noticeable drops in sub-question performance despite relatively high accuracy on the main question. This discrepancy suggests that while MLLMs can generate reasonable answers for the main question, they often fail to maintain consistent reasoning across intermediate steps, limiting their overall reasoning capability. Open-source models, such as Qwen2.5-VL-7B, exhibit even lower performance on sub-questions, particularly in Q1 and Q3, further emphasizing the challenge these models face in chain-of-thought reasoning, where the ability to logically connect intermediate reasoning steps is crucial. 

\noindent \textbf{SurgCoT Enables CoT Reasoning} \quad
SurgCoT facilitates surgical CoT reasoning by decomposing tasks into clinical sub-questions (Q1→Q2→Q3) and progressively integrating knowledge (\textbf{KE}) and spatiotemporal clues (\textbf{FC}) to guide progressive, transparent inference. 
As shown in Table~\ref{tab:Q_results}, the introduction of Knowledge Enhancement (\textbf{KE}) yields substantial improvements in sub-question performance across diverse model architectures. For instance, LLaVA-Med-7B~\cite{li2023llava} demonstrates a notable accuracy increase in main Q from 68.15\% to 75.22\% when transitioning from \textbf{BL} to \textbf{KE}. The subsequent integration of spatiotemporal clues (\textbf{FC}) further elevates performance, particularly in stages requiring fine-grained temporal and spatial analysis (Q2, Q3), where models must ground their reasoning in concrete visual evidence.

Fig.~\ref{case} illustrates a typical case when \texttt{Knowledge} and \texttt{Clue} are gradually introduced, the model refines its reasoning process, correcting initial mistakes and arriving at the correct answer. In Fig.~\ref{case}(1), the model's response is based solely on clinical knowledge from pretraining, without effectively integrating the video content. Fig.~\ref{case}(2) expands the model's reasoning process, but significant errors and deviations remain. Fig.~\ref{case}(3) shows that after adding \texttt{Knowledge}, the model adjusts its reasoning to produce a more accurate answer. Finally, 
Fig.~\ref{case}(4) shows that added \texttt{Clues} further improve performance, highlighting the benefits of SurgCoT for enhancing progressive reasoning.

%% file: sec/5_conclusion.tex
\section{Conclusion}
SurgCoT introduces a benchmark for evaluating surgical reasoning of MLLMs via a chain-of-thought framework, with multi-specialty fine-grained reasoning tasks. Evaluation of leading MLLMs shows significant gaps in progressive reasoning, \ie, often answering correctly with wrong intermediate reasoning steps. SurgCoT not only exposes the limitations but also provides a scaffold for improvement with structured reasoning and spatiotemporal grounding.

% In summary, SurgCoT provides a unified, clinically grounded benchmark for evaluating surgical MLLMs under a chain-of-thought framework, coupling broad cross-specialty procedural coverage with five fine-grained spatiotemporal reasoning dimensions and a three-stage (\texttt{Q1→Q2→Q3}) five-tuple (\texttt{Question→Option→Knowledge→Clue→Answer}) annotation protocol. Our evaluation of 12 state-of-the-art commercial, open-source, and medical MLLMs shows that, although commercial models currently lead in overall performance, all models still exhibit marked deficiencies in progressive spatiotemporal reasoning and causal judgment, often answering main questions correctly while failing on intermediate steps. At the same time, the observed gains under the progressive reasoning settings demonstrate that SurgCoT not only exposes these gaps but also offers a practical pathway toward improving model reasoning via structured CoT scaffolding and explicit spatiotemporal evidence, laying a reproducible foundation for future advances in surgery-aware MLLMs. 